%% file: INTERSPEECH2019_GUIDED_CTC.tex
\documentclass[a4paper]{article}

\usepackage{INTERSPEECH2019}

\usepackage{array}
\def\thline{\noalign{\hrule height 1pt}}

\usepackage{wrapfig}
\usepackage{multirow}
\usepackage{arydshln}

\usepackage{bm}

\usepackage{enumitem}
\setlist[description]{leftmargin=0.5\parindent,labelindent=0cm}

\usepackage{graphicx}
\usepackage{cite}

\newcommand{\figref}[1]{Figure \ref{#1}}
\newcommand{\tabref}[1]{Table \ref{#1}}

\newcommand{\secref}[1]{Section \ref{#1}}

\renewcommand{\columnsep}{5truemm}

\usepackage{subfigure}

\usepackage{cite}
\usepackage{xcolor}

\newcommand{\fii}[1]{\footnotesize {\it {#1}}}

\title{Guiding CTC Posterior Spike Timings\\for Improved Posterior Fusion and Knowledge Distillation}
\name{Gakuto Kurata$^\star$,~~Kartik Audhkhasi$^\dagger$
}
\address{
  $^\star$IBM Research - Tokyo\\
  $^\dagger$IBM T. J. Watson Research Center}
\email{gakuto@jp.ibm.com,~~kaudhkha@us.ibm.com}

\begin{document}

\maketitle

\begin{abstract}  
Conventional automatic speech recognition (ASR) systems trained from frame-level alignments can easily leverage posterior fusion to improve ASR accuracy and build a better single model with knowledge distillation.
End-to-end ASR systems trained using the Connectionist Temporal Classification (CTC) loss do not require frame-level alignment and hence simplify model training.
However, sparse and arbitrary posterior spike timings from CTC models pose a new set of challenges in posterior fusion from multiple models and knowledge distillation between CTC models.
We propose a method to train a CTC model so that its spike timings are guided to align with those of a pre-trained {\it guiding} CTC model.
As a result, all models that share the same guiding model have aligned spike timings.
 We show the advantage of our method in various scenarios including posterior fusion of CTC models and knowledge distillation between CTC models with different architectures.
 With the 300-hour Switchboard training data, the single word CTC model distilled from multiple models improved the word error rates to 13.7\%/23.1\% from 14.9\%/24.1\% on the Hub5 2000 Switchboard/CallHome test sets without using any data augmentation, language model, or complex decoder.
\end{abstract}


\section{Introduction}
\label{sec:introduction}
End-to-end (E2E) automatic speech recognition (ASR) using the Connectionist Temporal Classification (CTC) loss function has been gathering interest since it significantly simplifies model training pipelines\footnote{We do not focus on other major approaches to E2E ASR in this paper, such as attention-based encoder-decoder systems~\cite{bahdanau2016end} and joint CTC/sequence-to-sequence systems~\cite{hori2017joint}.}\cite{graves2006connectionist,miao2015eesen,sak2015learning,pundak16:_lower,miao2016empirical,zweig2017advances,Soltau2017}.
Prior to E2E ASR systems, the conventional training pipeline for Gaussian Mixture Model-Hidden Markov Model (HMM) systems~\cite{rabiner1989tutorial} and Deep Neural Network (DNN)-HMM hybrid systems~\cite{hinton2012deep} required output symbols for every input acoustic frame, that is, {\it frame-level alignment}, which made the training process complex and time-consuming.
Instead, E2E ASR only requires pairs of input feature sequences and output symbol sequences~\cite{graves2006connectionist,graves2013speech,graves2013generating,graves2014towards,hannun2014deep,hannun2017sequence}, such as phones~\cite{chorowski2015attention}, characters~\cite{chan2016listen}, words~\cite{Audhkhasi2017,Soltau2017}, or their combinations~\cite{liu2017gram,audhkhasi2017building}.

One advantage of training ASR models with frame-level alignments is that frame-level posterior fusion of multiple systems is easy.
Models trained from the same frame-level alignment data have the same target output symbol for each frame.
Thus, we can compute the average or a weighted average of the posteriors from multiple systems to yield better posteriors, which results in improved ASR accuracy~\cite{george17:_englis_conver_telep_speec_recog_human_machin} because such posterior averaging effectively creates a mixture model comprising of the different systems being combined.
We can also use the (weighted) average of posteriors from multiple acoustic models as a teacher to train a single student model via knowledge distillation\footnote{Knowledge distillation is also called teacher-student training, where a {\it student} model is distilled from a {\it teacher} model.}~\cite{ba2014deep,li2014learning,hinton2015distilling,chebotar2016distilling,fukuda2017efficient}.

In contrast, as noted in prior literature~\cite{graves2006connectionist,sak2015learning}, CTC models emit very {\it spiky} posterior distributions\footnote{We call the highest posterior except for blank and silence at each time index a ``spike'' hereafter.} where most frames emit the garbage {\it blank} symbol with high probability and only a few frames emit the target symbols of interest.
As a result, CTC models trained with the same training data can have different spike timings (typical examples are shown later in \figref{fig:uni_phone} and \ref{fig:bi_word}), which poses a new set of challenges.
Due to non-aligned spike timings, we cannot make better posteriors by computing their (weighted) average across multiple systems~\cite{sak2015acoustic}.
Hence, it's difficult to improve ASR accuracy by a na\"ive posterior fusion, and also to use the (weighted) average of posteriors as a teacher to train a single model via knowledge distillation.

Disagreement of CTC spike timings has been discussed in the context of knowledge distillation from bidirectional LSTM (BiLSTM) CTC to unidirectional LSTM (UniLSTM) CTC models~\cite{kurata18:_improv_knowl_distil_from_bi} and from multiple accent-specific models to a single multi-accent model~\cite{ghorbani2018advancing}.
\cite{kurata18:_improv_knowl_distil_from_bi} proposed searching for a similar posterior from a BiLSTM CTC model at a different time index to train a UniLSTM model.
This approach only focused on knowledge distillation from BiLSTM to UniLSTM and cannot be simply extended to a posterior fusion of multiple systems.
\cite{ghorbani2018advancing} attempted to train an accent-specific model for each accent by feeding training data of the target accent to a common teacher model trained from the data of all accents via knowledge distillation.
As a result, each accent-specific model has aligned spike timings with the common teacher model.
This method did not address knowledge distillation between different neural network (NN) architectures, such as between BiLSTM and UniLSTM CTC models.

In this paper, we propose a method to explicitly guide the CTC spike timings to be aligned with those from a pre-trained CTC model with the NN architecture of interest.
We call this pre-trained CTC model a {\it guiding model}.
More specifically, when training a CTC model, in addition to a normal CTC loss, we add a new loss term that forces the spikes from the model being trained to occur at the same time as those from the guiding model.
Hence, models guided by the same guiding model have aligned spike timings.
The advantages of our proposed method are as follows:
 \begin{description}
  \item[Posterior fusion of multiple CTC models]~\\
	     We can make better posteriors by averaging aligned posteriors from multiple models that share the same guiding model and improve ASR accuracy as a result.
	     In addition, we can use these averaged posteriors as a teacher to train a single model with improved ASR accuracy through knowledge distillation.
 \item[{\parbox[b]{\columnwidth}{Knowledge distillation between CTC models with arbitrary neural network architectures}}]
	     The spike timings of CTC models with different NN architectures can be significantly different, which makes knowledge distillation between such models difficult.
	     A typical case is knowledge distillation from a BiLSTM CTC model to a UniLSTM CTC model.
	     By training a teacher model using a guiding model that has the same architecture as the final desired student model, the trained teacher model can have the spike timings appropriate for the student model and can be used in knowledge distillation.
  \item[Improved speech recognition accuracy]~\\
	     The CTC model guided by the guiding model has improved ASR accuracy compared to the model without the guiding model.
	     The guiding model promotes a good alignment path and hence does not allow the CTC training to assign sufficient probabilities to bad alignment paths, which results in an improved accuracy.
 \end{description}

 We will show these advantages in various experiments with the standard English Switchboard conversational telephone speech data.

\section{Connectionist Temporal Classification}
Let $\bm{y}$ denote a length-$L$ sequence of target output symbols.
Let $\bm{X}$ denote acoustic feature vectors over $T$ time steps.
The conventional alignment-based DNN/HMM hybrid system training requires $L$ to be equal to $T$~\cite{hinton2012deep}.
The alignment-free CTC introduces an extra blank symbol $\phi$ that expands the length-${L}$ sequence $\bm{y}$ to a set of length-${T}$ sequences $\Phi(\bm{y})$.
Each sequence $\bm{{\hat y}} \in \Phi(\bm{y})$ is one of the {\it CTC alignments} between $\bm{X}$ and $\bm{y}$.
The CTC loss is defined as the summation of symbol posterior probabilities over all possible CTC alignments: ${\cal L}_{CTC} = - \sum_{\bm{{\hat y}} \in \Phi(\bm{y})} P(\bm{{\hat y}} | \bm{X}).$
We use phones and words as target output symbols, namely {\it phone CTC} and {\it word CTC}, for evaluation in \secref{sec:experiments}.

\input{FIGURE_GUIDED_CTC_TRAINING}

\input{FIGURE_EXPERIMENTS}

\section{Guided CTC training}
\label{sec:guided-ctc-training}
In order to guide the spike timings of CTC models, we propose a two-step training process and name it {\it guided CTC training}.
First, we train a guiding CTC model that has appropriate spike timings for the final target use case.
Then, we train another CTC model (the {\it guided model}) so that similar spike timings as that of the guiding model can be obtained.
Note that as long as the set of the output symbols is the same, different NN architectures can be used for the guiding and the guided models.

\figref{fig:guided_ctc_training} shows a schematic diagram on how to guide the spike timings of the guided model by using the pre-trained guiding CTC model, assuming that we have completed the training of the guiding model in advance.
When training the guided CTC model, for a training sample $\bm{X}$ of $T$ time steps, we feed it to the guiding CTC model, do the forward pass, and obtain posteriors for each time index.
We convert these posteriors to a mask $M(\bm{X})$ by setting a 1 at the output symbol with the highest posterior and 0 at other symbols at each time index.
In cases where the blank symbol $\phi$ has the highest posterior, we set 0 for all symbols at this time index.
This mask passes the output symbol that the guiding CTC model emits at each time index.
Then we feed the same training sample to the guided CTC model being trained and obtain posteriors $P(\bm{X})$.
By the Hadamard (element-wise) product of the mask and the posteriors, we obtain the masked posteriors $\hat{P}(\bm{X}) = M(\bm{X}) \circ P(\bm{X})$.
The summation of masked posteriors $\hat{P}(\bm{X})$ becomes greater if the guided CTC model being trained has spikes for the same output symbol at the same times as the guiding CTC model.
Thus, by maximizing this summation, we can guide the spike timings of the guided CTC model to be the same as those of the guiding CTC model.
We call the summation of the masked posteriors multiplied by $-1$ the {\it guide loss} defined as ${\cal L}_{G} = -1 \cdot \sum \hat{P}(\bm{X})$~\footnote{The logarithmic guide loss is equivalent to a frame-level cross-entropy where the target is a sequence of the output symbols with the highest posterior from the guiding model. Note that the mask $M(\bm{X})$ is derived from this target sequence.}.
The overall loss for the guided CTC training becomes the summation of the standard CTC loss  and the guide loss as ${\cal L} = {\cal L}_{CTC} + {\cal L}_{G}$.

\section{Experiments}
\label{sec:experiments}
We conducted ASR experiments on phone and word CTC models to determine the advantages of our proposed guided CTC training.
We used the standard 300-hour Switchboard English conversational telephone speech as training data~\cite{george17:_englis_conver_telep_speec_recog_human_machin}.
All models were trained for 20 epochs using stochastic gradient descent with the Nesterov momentum of 0.9 and a learning rate starting from 0.03 and annealing at $\sqrt{0.5}$ per-epoch after the 10th epoch.
We set the batch size to 128 for phone CTC models and 96 for word CTC models.

We trained multiple models for each experiment to investigate the effect of posterior fusion.
By using different seeds for random parameter initialization, even models with exactly the same architecture trained with the same procedure can have sufficient diversity to benefit from posterior fusion~\cite{hinton2015distilling}.
We follow this procedure and also randomized the order of the training data after the first epoch (SortaGrad)~\cite{amodei2016deep}.

\input{TABLE_UNI_PHONE}

\input{FIGURE_POSTERIORS_ALL}

\subsection{Posterior fusion of UniLSTM phone CTC models}
\label{sec:post-fusi-unilstm}
First, we conducted experiments on posterior fusion with UniLSTM phone CTC models.
We used 40-dimensional logMel filterbank energies, their delta, and double-delta coefficients with frame stacking and skipping rate of 2~\cite{sak2015fast}, resulting in 240-dimensional features.
We used 44 phones from the Switchboard pronunciation lexicon~\cite{george17:_englis_conver_telep_speec_recog_human_machin} and the blank symbol.
For decoding, we trained a 4-gram language model with 24M words from the Switchboard + Fisher transcripts with a vocabulary size of 30K.
We constructed a CTC decoding graph similar to the one in \cite{miao2015eesen}.
For NN architecture, we stacked 6 UniLSTM layers with 640 units and a fully-connected linear layer of 640$\times$45, followed by a softmax activation function.
All NN parameters were initialized to samples of a uniform distribution over $(-\epsilon,\epsilon)$, where $\epsilon$ is the inverse square root of the input vector size.
For evaluation, we used the Hub5-2000 Switchboard (SWB) and CallHome (CH) test sets.

\figref{fig:experiments} shows the flow of experiments and \tabref{tab:uni_phone_ctc} lists the results.
We trained 4 UniLSTM models with the standard training.
In {\tt 1A}, we averaged the Word Error Rates (WERs) for the 4 decoding outputs from the 4 models.
For posterior fusion in {\tt 1B}, we averaged the posteriors from the 4 models and used it for decoding with the graph.
For comparison, we also combined the 4 decoding outputs from the 4 models by using the Recognizer Output Voting Error Reduction (ROVER)~\cite{fiscus1997post,wang2017residual,fritz2017simplified}, as shown in {\tt 1C}.
Comparing {\tt 1A} and {\tt 1B}, we can see that since the posteriors from the 4 models were not aligned, we did not benefit from posterior fusion.
ROVER gave us a solid improvement in {\tt 1C}.

For the proposed guided CTC training, as shown in the right of \figref{fig:experiments}, we first trained a UniLSTM phone CTC model with the same architecture and training data and then used it as a guiding model to train 4 guided models.
{\tt 1D} indicates the average WERs of the 4 decoding outputs by the 4 guided UniLSTM models trained by the proposed guided CTC training.
Comparing {\tt 1A} and {\tt 1D}, we can see a WER reduction by the proposed guided CTC training.
The guide loss promotes a good CTC alignment path and assigns lower probabilities to bad CTC alignment paths on the basis of the guiding model during the CTC training\footnote{An na\"ive alternative is to conduct the standard UniLSTM CTC training starting from a pre-trained guiding CTC model. By this approach, we obtained the WERs of 15.0\% and 27.3\%. The larger improvements obtained in {\tt 1D} demonstrate the advantage of our proposed guided CTC training.}.
{\tt 1E} and {\tt 1F} show WERs by the posterior fusion and ROVER of the 4 models trained with the proposed guided CTC training.
Due to the aligned posteriors, we benefit from the posterior fusion that outperformed the ROVER.
Note that posterior fusion requires decoding with the graph just once, while ROVER needs 4 separate decodes.
This is another advantage of posterior fusion realized by the proposed guided training.

\figref{fig:uni_phone} shows posteriors of multiple UniLSTM phone CTC models for the same utterance, ``this (DH IH S) is (IH S) true (T R UW)'', in the SWB test set, where symbols in parentheses represent phones.
As expected, phone spike positions were not aligned between the UniLSTM CTC models.
\figref{fig:uni_phone_guided} shows posteriors of the guiding UniLSTM CTC model at the top and the UniLSTM CTC models guided by the guiding model at the top with the proposed training.
The spikes from the guided models were temporally smoothed and the spike positions of the guiding models were covered by the smoothed spikes of the guided models.
Hence, the spike timings from multiple guided models overlapped, which underpins the improved WERs by posterior fusion.

\input{TABLE_BI_TO_UNI_PHONE}

\input{TABLE_BI_WORD_ALL}

\input{TABLE_COMPARISON}

\subsection{Knowledge distillation from BiLSTM phone CTC to UniLSTM phone CTC models}
Next, we explored knowledge distillation from BiLSTM to UniLSTM phone CTC models.
UniLSTM CTC models are more suitable for deployment in actual streaming ASR services and products~\cite{sak2015acoustic,soltau17:_reduc} and closing the ASR accuracy gap between UniLSTM and BiLSTM CTC models is critical~\cite{kim18:_improv,takashima18:_ctc,kurata18:_improv_knowl_distil_from_bi}.
Here, we used the same speaker-independent (SI) 240-dimensional feature vectors  as the previous experiments while keeping the actual use-case of streaming ASR in mind.
For teacher BiLSTM phone CTC models, we stacked 6 BiLSTM layers with 320 units each in the forward and backward layers and a fully-connected linear layer of 640$\times$45, followed by a softmax activation function.
For UniLSTM phone CTC models. we used the same architecture as in the previous experiments.
All NN parameters were initialized in the same fashion as the previous experiments.
For evaluation, we used the same SWB and CH test sets.

\tabref{tab:bi_to_uni_phone} shows the results.
{\tt 2A} and {\tt 2B} indicate the WERs by UniLSTM and BiLSTM phone CTC models trained with the standard procedure.
As a na\"ive approach, in {\tt 2C}, we trained a UniLSTM model through knowledge distillation from BiLSTM models with minimizing the frame-wise KL divergence, but saw degradation in WERs.
The spike timings from BiLSTM were not appropriate for UniLSTM, as also demonstrated in \figref{fig:bi_and_uni_phone}, where the spike timings for the same utterance were completely different between the UniLSTM and BiLSTM models.
In addition, posteriors from BiLSTM models trained with the standard training were not aligned and thus knowledge distillation from their posterior fusion was not successful.

{\tt 2D} is the key step where we first trained a guiding UniLSTM model and then trained a BiLSTM model guided by the guiding UniLSTM model.
The trained BiLSTM model had spike timings appropriate for UniLSTM model as shown in \figref{fig:bi_to_uni_phone} while posteriors were estimated with bidirectional context.
Comparing {\tt 2B} and {\tt 2D}, we can see that {\tt 2D} had a worse WER because the spike positions were unnatural for a BiLSTM.
However, this model can serve as an appropriate teacher to train a UniLSTM model.
We trained multiple BiLSTM models guided by the same guiding UniLSTM model and used their posterior fusion as a teacher to train a single student UniLSTM model by knowledge distillation, where the frame-wise KL divergence from the student model to the fused teacher posterior was minimized.
{\tt 2E} shows the WERs by the UniLSTM models distilled from the posterior fusion of 1, 4, and 8 BiLSTM models guided by the same guiding UniLSTM model\footnote{``Posterior fusion of 1 guided model'' is used for better readability and indicates the use of posteriors from the 1 guided model directly.}.
Even in the case of just 1 guided BiLSTM model, the WER for SWB was reduced to 13.4\%, which was equivalent to reducing the gap between BiLSTM and UniLSTM by 54.3\%\footnote{We reported a smaller accuracy gap reduction (45.2\%) with a similar setting in our former paper~\cite{kurata18:_improv_knowl_distil_from_bi}. Besides, this method could not be extended to use a posterior fusion as a teacher.}.
By increasing the number of BiLSTM models, the WER for SWB was reduced to 12.9\%, which equals a 68.6\% accuracy gap reduction.
As in the bottom two posteriors in \figref{fig:bi_to_uni_phone}, the temporally smoothed spikes of the guided BiLSTM models overlapped around the spikes of the guiding UniLSTM model, which underpins the improved WERs by increasing the number of guided BiLSTM models.

\subsection{Posterior fusion and knowledge distillation of BiLSTM word CTC models}
We applied our proposed guided CTC training to BiLSTM word CTC models.
For input acoustic features, we added 100-dimensional i-vectors for each speaker extracted in advance and appended them to the same SI feature used in the previous experiments, resulting in 340-dimensional feature vectors~\cite{dehak2011front}.
We selected words with at least 5 occurrences in the training data~\cite{Audhkhasi2017,audhkhasi2017building}.
This resulted in an output layer with 10,175 words and the blank symbol.
We stacked 6 BiLSTM layers with 512 units each in the forward and backward layers (BiLSTM encoder), added 1 fully-connected linear layer with 256 units to reduce computation~\cite{sainath2013low}, and put 1 fully-connected linear layer of 256$\times$10,176, followed by a softmax activation function.
For better convergence, we initialized the BiLSTM encoder part with the trained BiLSTM phone CTC model\footnote{To initialize multiple word CTC models for posterior fusion and ROVER, we trained multiple phone CTC models with different parameter initialization and training data order.}~\cite{Audhkhasi2017,audhkhasi2017building}.
Other parameters were initialized in similar fashion as the phone CTC models.
For decoding, we performed a simple peak-picking over the output word posterior distribution, and removed repetitions and blank symbols. 
For evaluation, in addition to SWB and CH, we used RT02, RT03, and RT04 test sets~\cite{george17:_englis_conver_telep_speec_recog_human_machin}.

\tabref{tab:bi_word_ctc} shows the results.
We trained 4 BiLSTM models with the standard training and conducted posterior fusion and ROVER.
{\tt 3A} indicates the average WERs of 4 decoding outputs from the 4 models and {\tt 3B} and {\tt 3C} indicate the WERs by posterior fusion and ROVER, respectively.
Comparing {\tt 3A} and {\tt 3B}, since the posteriors from the 4 models were not aligned and the spikes from word models were much more sparse than those from phone models, we saw a significant degradation in accuracy.
Due to the sparse non-aligned word spikes, ROVER also did not improve the ASR accuracy, as in {\tt 3C}.
\figref{fig:bi_word} shows the word posteriors from BiLSTM word CTC models that were sparse and non-aligned, which caused degradation in {\tt 3B} and {\tt 3C}.

For the proposed guided CTC training, we first trained a BiLSTM word CTC model with the same architecture and training data and then used it as a guiding model to train the 4 guided models.
{\tt 3D} indicates the average WERs of 4 decoding outputs by the 4 BiLSTM word models trained by the proposed guided CTC training.
Comparing {\tt 3A} and {\tt 3D}, we can see WER reduction by the proposed guided CTC training, the same as with the case of phone CTC models in \secref{sec:post-fusi-unilstm}.
{\tt 3E} and {\tt 3F} show WERs by the posterior fusion and ROVER of the 4 guided models.
Due to the aligned posteriors obtained by the guided CTC training, both posterior fusion and ROVER improved ASR accuracy compared with {\tt 3D} while posterior fusion outperformed ROVER.
Finally, we used the posterior fusion from the guided BiLSTM models as a teacher and trained a single student BiLSTM word CTC model by knowledge distillation with minimizing the frame-wise KL divergence.
{\tt 3G} shows the results where the WERs were further decreased from {\tt 3D} consistently over all test sets.

\tabref{tab:comparison} shows a comparison on WERs with the published CTC-based direct acoustic-to-word models trained from the standard 300-hour Switchboard data without using any data augmentation or language model.
To the best of our knowledge, our best single model trained by knowledge distillation outperformed the published CTC models trained in a purely E2E fashion in {\tt 4A} and {\tt 4B}~\cite{audhkhasi2017building,sanabria2018hierarchical}.
By combining CTC and cross-entropy (CE) training using the alignment between the acoustic frame and the labels, as in a conventional non-E2E system, 13.0\% and 23.4\%\footnote{By using data augmentation with speed perturbation, \cite{yu2018multistage} further improved the WERs to 11.4\% and 20.8\%.} were achieved in {\tt 4C}~\cite{yu2018multistage}.
Our best single model achieved the comparable WERs of 13.7\% and 23.1\% with purely E2E training without any frame-level alignments.

\input{TABLE_COVERAGE}

\subsection{Analysis on spike timings}
Looking at the posteriors in \figref{fig:posteriors}, we can see that spikes from models that were trained with the standard training (we call them {\it non-guided} models) were not aligned, while the spikes from the guiding and the guided models and those of the guided models that share the same guiding model were aligned.
We quantitatively analyzed the coverage ratio by investigating if the posterior spikes (except for the blank and silence) from one model were covered by the spikes from the other model at the same time index.
We used the training data and the test data (SWB) for this analysis; the results are shown in \tabref{tab:coverage}.
Note that in the case of ``guiding and guided models'', we investigated if the spikes from the guiding model were covered by the corresponding guided model.
For other cases, we randomly picked 2 models trained using the same training procedure with changing the parameter initialization and the training data order.
For both UniLSTM phone CTC and BiLSTM word CTC models, the coverage ratio of 2 non-guided models was low.
Especially, due to sparse spikes, the coverage ratio of BiLSTM word CTC was much lower, which underpins the poor WERs by the posterior fusion in {\tt 3B} of \tabref{tab:bi_word_ctc}.
The coverage ratio between the guiding and guided model was improved and the ratio between 2 guided models was slightly worse, but much higher than the ratio between the 2 non-guided models.
Comparing the training and test data, the training data had a slightly higher coverage ratio, but the same trends were seen for all combinations of models, which supports the experimental results.

\section{Conclusion}
\label{sec:conclusion}
We proposed a method to guide spike timings of CTC models by using a pre-trained CTC model with the NN architecture of interest.
We demonstrated its advantages in various scenarios including posterior fusion of multiple CTC models and knowledge distillation between CTC models.
Through the experiments, we achieved state-of-the-art WERs in the CTC-based direct acoustic-to-word setting without using any data augmentation or language model.
By qualitatively and quantitatively investigating the posterior spike timings from the CTC models trained with the proposed guided CTC training, we confirmed that the spike timings were aligned between the guiding and the guided models and between the multiple guided models.



\end{document}

%% file: FIGURE_GUIDED_CTC_TRAINING.tex
\begin{figure}[t]
  \begin{center}
   \includegraphics[width=1.0\columnwidth]{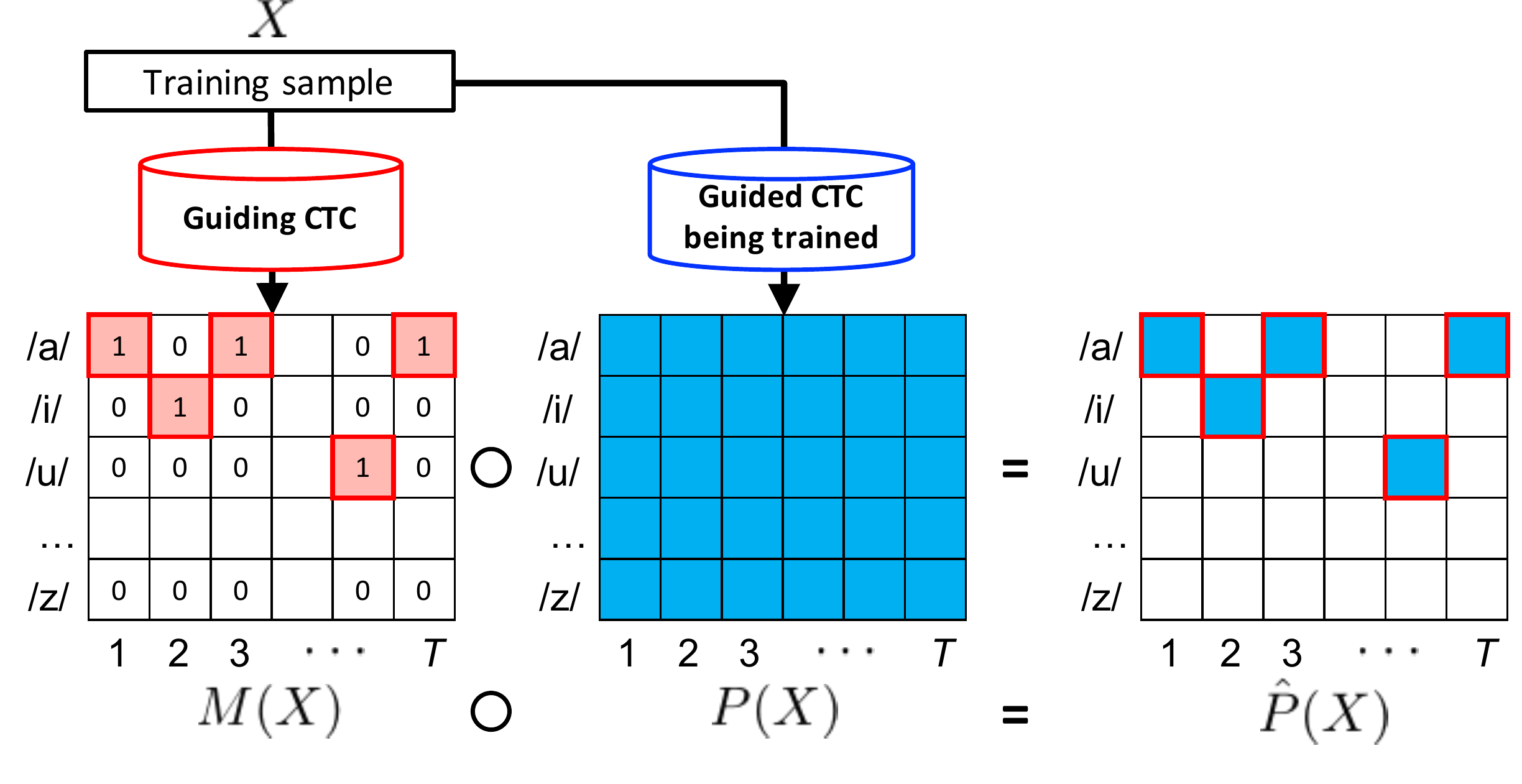}
   \vspace{-5mm}
   \caption{Schematic diagram of proposed guided CTC training. Vertical and horizontal axes of the matrices indicate output symbols and time indexes, respectively. The guiding CTC model at the left makes a mask that sets a 1 only at the output symbol of the highest posterior at each time index. This mask is applied to the posterior from the model being trained by Hadamard (element-wise) product. The summation of the masked posteriors at the right is multiplied by minus one and is minimized jointly with the CTC loss.}
  \label{fig:guided_ctc_training}
  \end{center}
	\end{figure}

%% file: FIGURE_EXPERIMENTS.tex
\begin{figure}[t]
  \begin{center}
   \includegraphics[width=1.0\columnwidth]{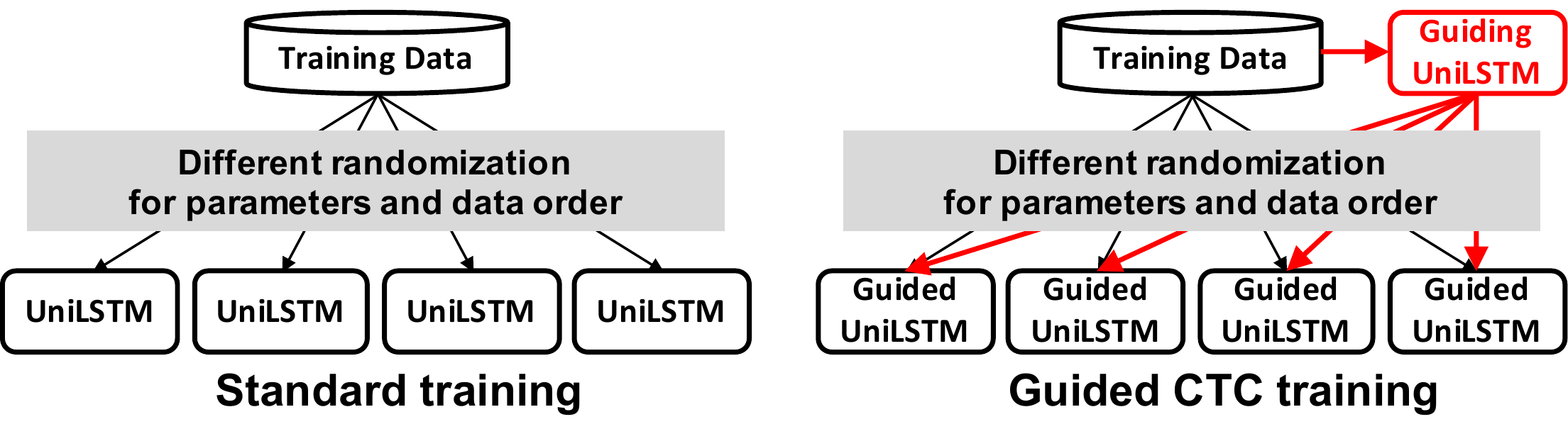}
   \vspace{-5mm}
   \caption{Standard and guided CTC training for UniLSTM phone CTC models.}
  \label{fig:experiments}
  \end{center}
	\end{figure}

%% file: TABLE_UNI_PHONE.tex
\begin{table}[t]
 \begin{center}
  \caption{WERs for posterior fusion and ROVER of UniLSTM phone CTC models. [\%]}
  \vspace{-3mm}
  \label{tab:uni_phone_ctc}
   \begin{tabular*}{\columnwidth}{@{\extracolsep{\fill}}lcc}
    \thline
    ~ & SWB & CH \\
    \thline
    {\tt 1A~}UniLSTM & 15.3 & 27.6 \\
    {\tt 1B~}4$\times$ posterior fusion of {\tt 1A~} & 15.4 & 28.8 \\
    {\tt 1C~}4$\times$ ROVER of {\tt 1A~}& 14.1 & 26.1 \\
    \hline
    {\tt 1D~}UniLSTM guided by UniLSTM& 14.4 & 26.2 \\
    {\tt 1E~}4$\times$ posterior fusion of {\tt 1D~}& 12.9 & 24.2 \\
    {\tt 1F~}4$\times$ ROVER of {\tt 1D~}& 13.7 & 24.5 \\
    \thline
   \end{tabular*}
 \end{center}
 \vspace{-3mm}
\end{table}

%% file: FIGURE_POSTERIORS_ALL.tex
 \begin{figure*}
  \def\subfigcapskip{-3pt}
  \begin{center}
   \begin{tabular}{c}

    \begin{minipage}{0.31\textwidth}
     \centering\subfigure[\fii{UniLSTM phone CTC models.}]{\includegraphics[width=1.0\hsize]{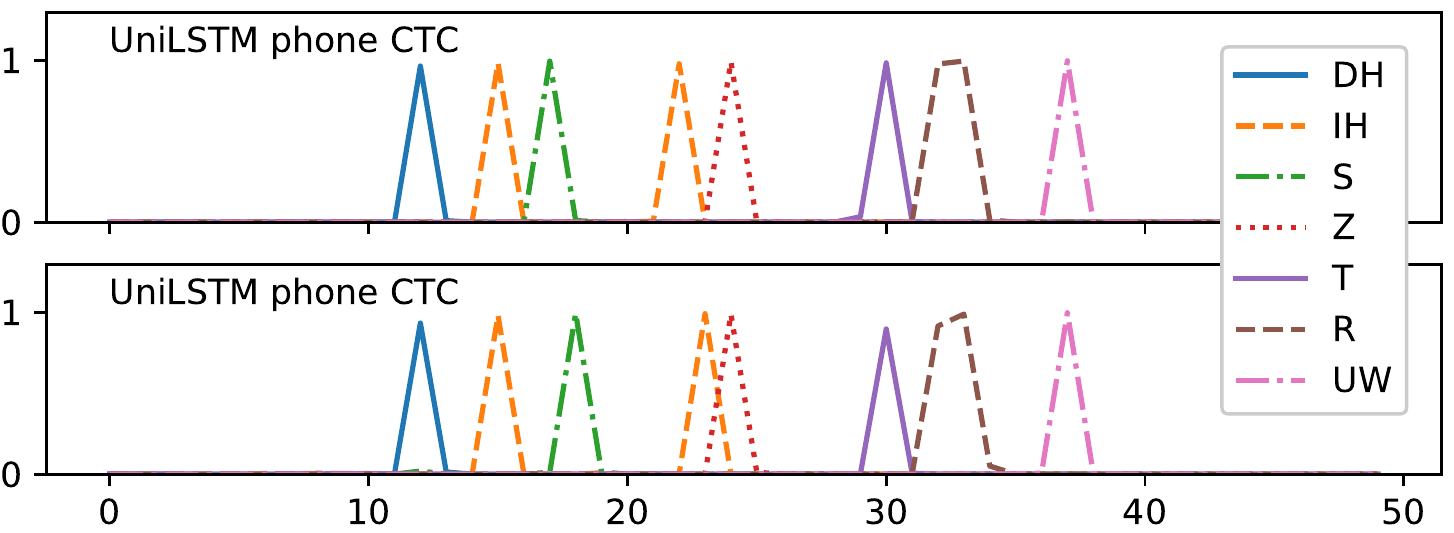}\label{fig:uni_phone}}
    \end{minipage}

    \begin{minipage}{0.02\textwidth}
      ~
    \end{minipage}  

    \begin{minipage}{0.31\textwidth}
     \setcounter{subfigure}{2}
     \centering\subfigure[\fii{UniLSTM and BiLSTM phone CTC models.}]{\includegraphics[width=1.0\hsize]{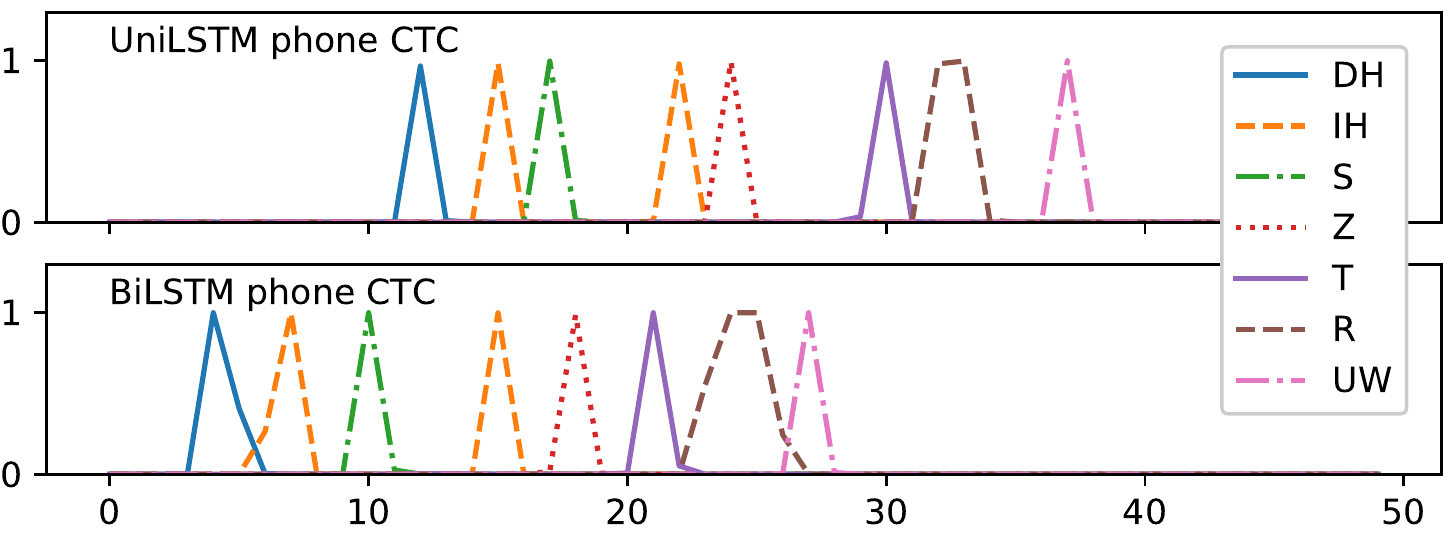}\label{fig:bi_and_uni_phone}}
    \end{minipage}

    \begin{minipage}{0.02\textwidth}
      ~
    \end{minipage}  

     \begin{minipage}{0.31\textwidth}
      \setcounter{subfigure}{4}
     \centering\subfigure[\fii{BiLSTM word CTC models.}]{\includegraphics[width=1.0\hsize]{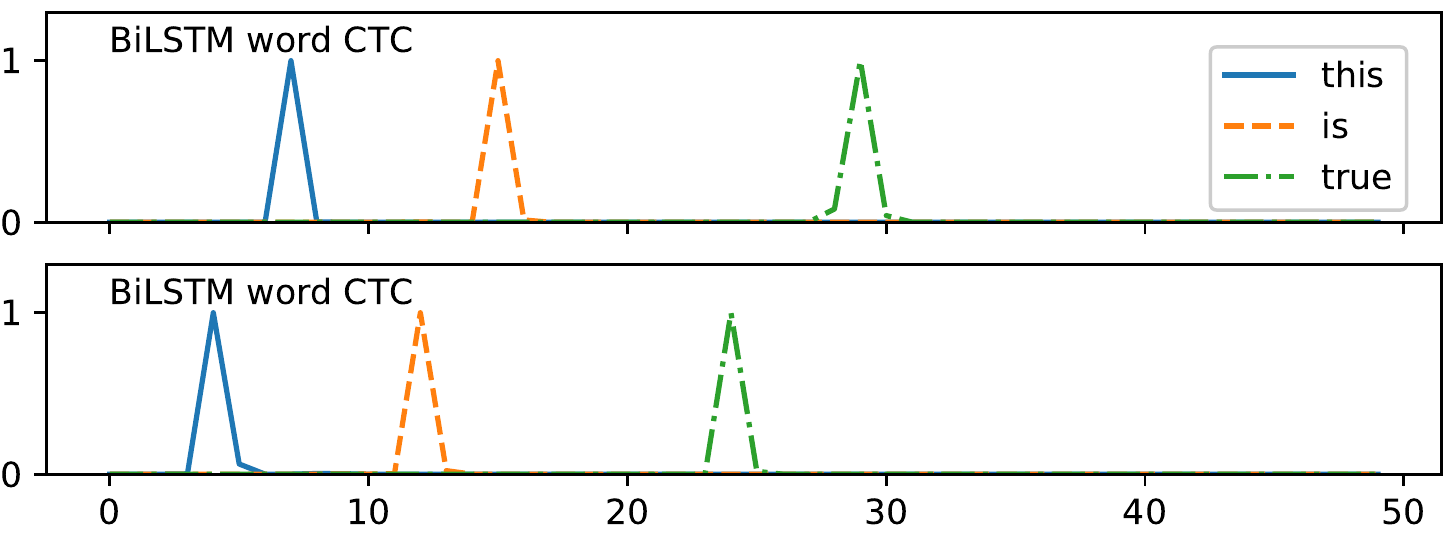}\label{fig:bi_word}}
    \end{minipage}    

    \\
    
     \begin{minipage}{0.31\textwidth}
                \setcounter{subfigure}{1}
      \centering\subfigure[\fii{UniLSTM phone CTC models. Bottom two models are guided by the top model.~~~~~~~~~~~~~~~~~~~~~~~~~~~~~~~~}]{\includegraphics[width=1.0\hsize]{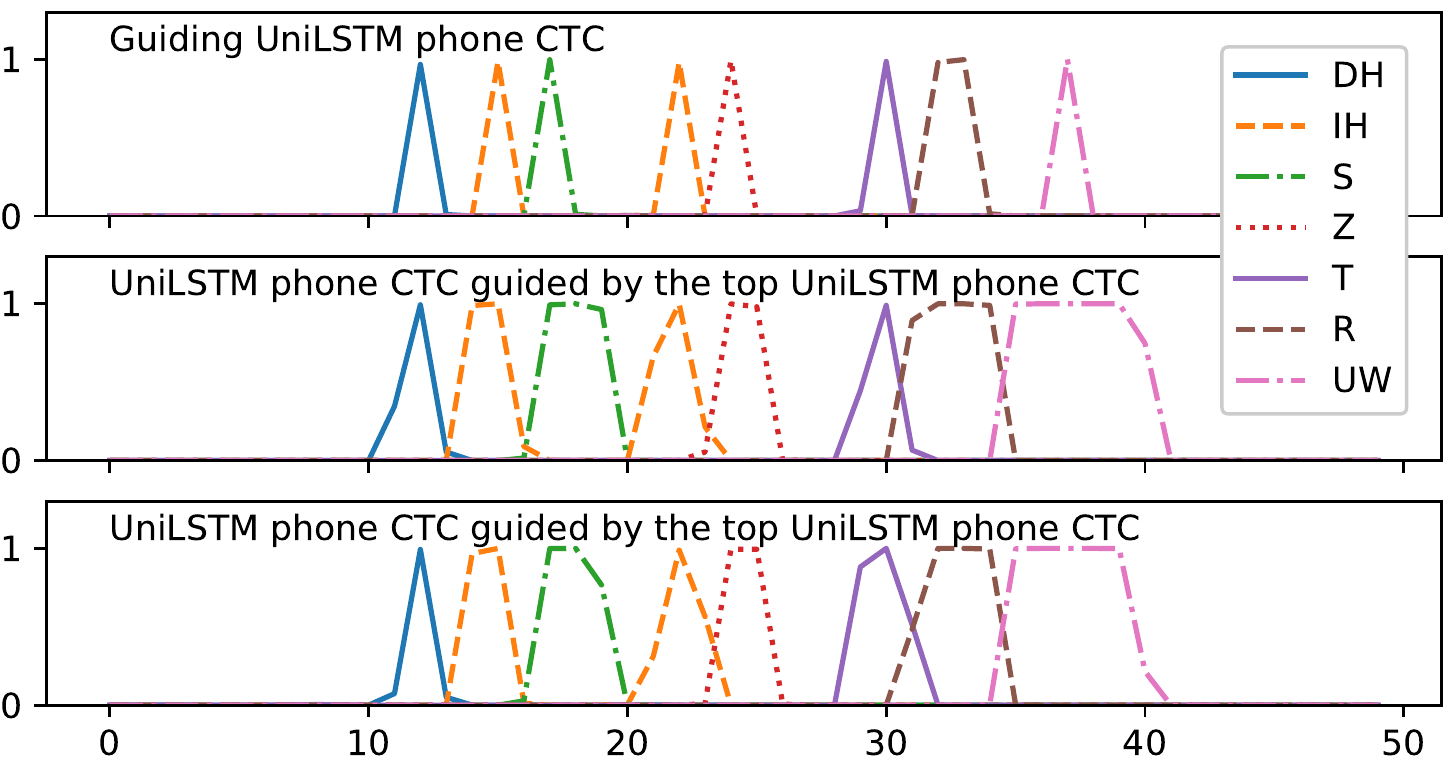}\label{fig:uni_phone_guided}}
    \end{minipage}    

    \begin{minipage}{0.02\textwidth}
      ~
    \end{minipage}  
    
     \begin{minipage}{0.31\textwidth}
                      \setcounter{subfigure}{3}
      \centering\subfigure[\fii{UniLSTM and BiLSTM phone CTC models. Bottom two BiLSTM models are guided by the top UniLSTM model.}]{\includegraphics[width=1.0\hsize]{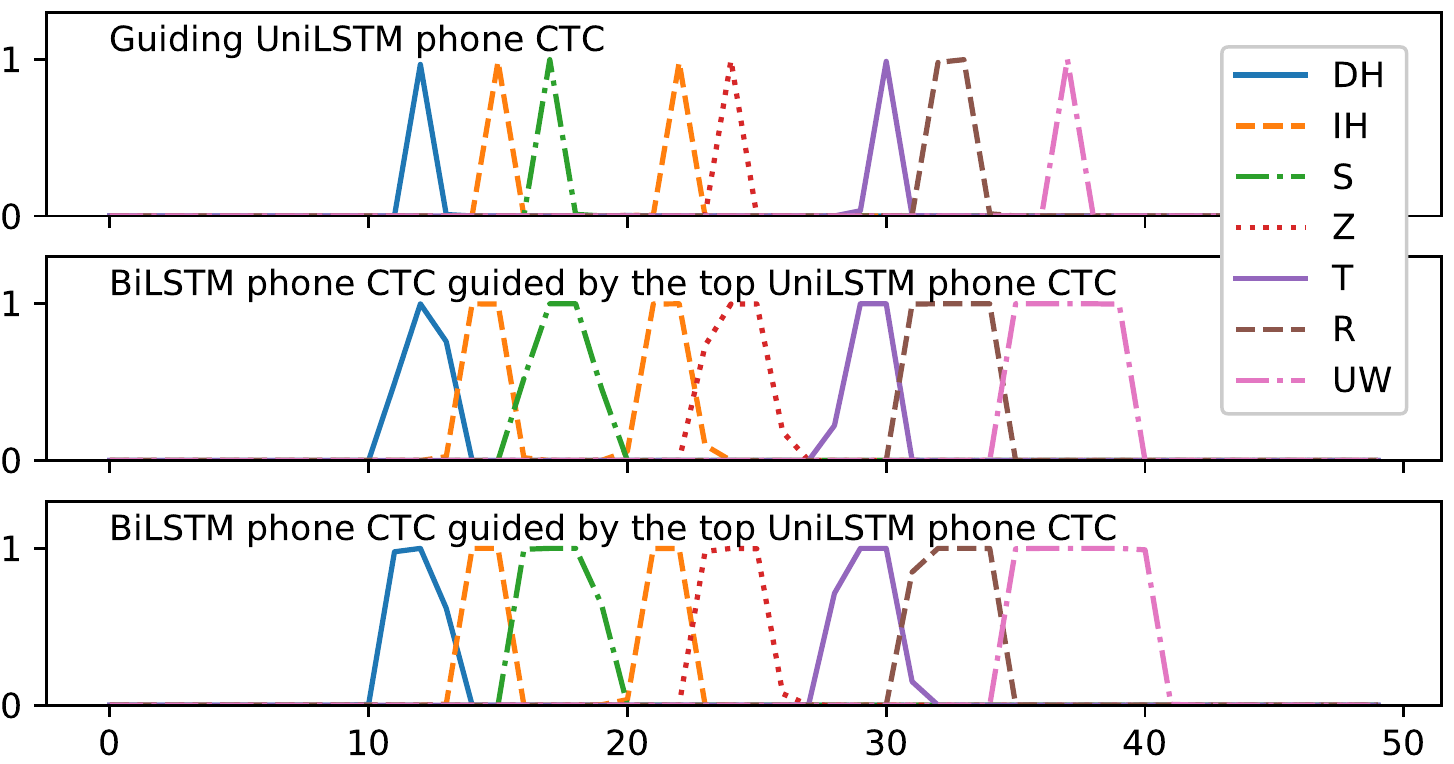}\label{fig:bi_to_uni_phone}}
    \end{minipage}    

    \begin{minipage}{0.02\textwidth}
      ~
    \end{minipage}  
    
     \begin{minipage}{0.31\textwidth}
      \setcounter{subfigure}{5}
      \centering\subfigure[\fii{BiLSTM word CTC models. Bottom two models are guided by the top model.~~~~~~~~~~~~~~~~~~~~~~~~~~~~~~~~~~~~~~~~~~}]{\includegraphics[width=1.0\hsize]{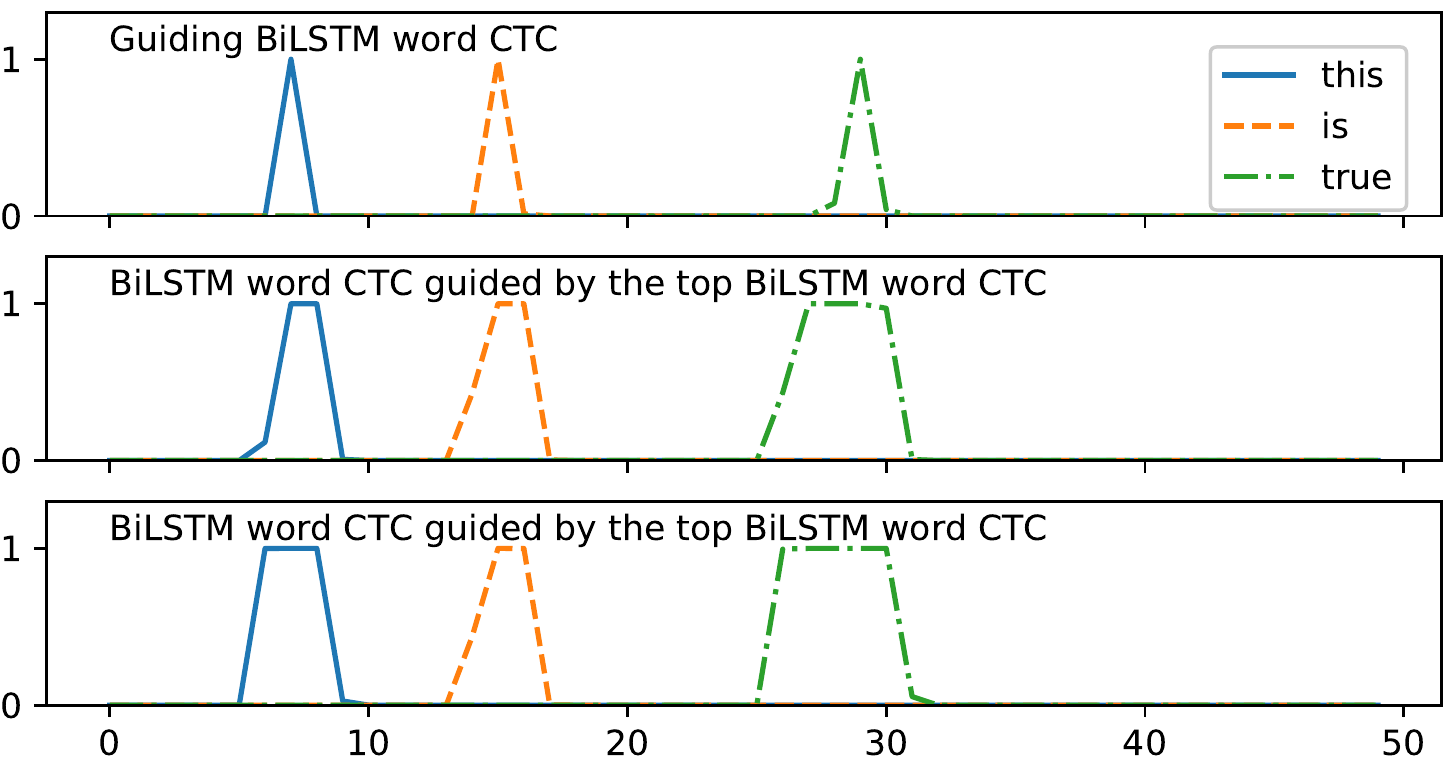}\label{fig:bi_word_guided}}
    \end{minipage}

   \end{tabular}
  \end{center}
  \vspace{-5mm}
   \caption{Posteriors for ``this (DH IH S) is (IH S) true (T R UW)'' in the SWB test set.}
 \label{fig:posteriors}

 \end{figure*}

%% file: TABLE_BI_TO_UNI_PHONE.tex
\begin{table}[t]
 \begin{center}
  \caption{WERs for knowledge distillation from BiLSTM to UniLSTM phone CTC models. [\%]}
  \vspace{-3mm}
  \label{tab:bi_to_uni_phone}
   \begin{tabular*}{\columnwidth}{@{\extracolsep{\fill}}lcc}
    \thline
    ~ & SWB & CH \\
    \thline
    {\tt 2A~}UniLSTM & 15.3 & 27.6 \\
    {\tt 2B~}BiLSTM & 11.8 & 21.8 \\
    \hline
    {\tt 2C~}UniLSTM distilled from & ~ & ~ \\
    ~~~~~~~~~~1$\times$ BiLSTM ({\tt 2B})& 17.1 & 29.9 \\
    ~~~~~~~~~~4$\times$ BiLSTMs ({\tt 2B})& 29.4 & 32.7 \\
    \hline
    {\tt 2D~}BiLSTM guided by UniLSTM& 12.4 & 22.6 \\
    \hline
    {\tt 2E~}UniLSTM distilled from & ~ & ~ \\
    ~~~~~~~~~~1$\times$ BiLSTM guided by UniLSTM ({\tt 2D})& 13.4 & 25.4 \\
    ~~~~~~~~~~4$\times$ BiLSTMs guided by UniLSTM ({\tt 2D})& 12.9 & 24.8 \\
    ~~~~~~~~~~8$\times$ BiLSTMs guided by UniLSTM ({\tt 2D})& 12.9 & 24.7 \\    
    \thline
   \end{tabular*}
 \end{center}
 \vspace{-3mm}
\end{table}

%% file: TABLE_BI_WORD_ALL.tex
\begin{table*}
 \begin{center}
  \caption{WERs for posterior fusion, ROVER, and knowledge distillation of BiLSTM word CTC models. [\%]}
  \vspace{-3mm}
  \label{tab:bi_word_ctc}
   \begin{tabular*}{\textwidth}{@{\extracolsep{\fill}}lcccccc}
    \thline
    ~ & SWB & CH & RT02 & RT03 & RT04 & Avg. \\
    \thline
    {\tt 3A~}BiLSTM & 14.9 & 24.1 & 23.7 & 24.1 & 22.6 & 21.9 \\
    {\tt 3B~}4$\times$posterior fusion of {\tt 3A~}& 48.2 & 57.7 & 57.7 & 58.9 & 59.3 & 56.4 \\
    {\tt 3C~}4$\times$ROVER of {\tt 3A~}& 16.0 & 23.2 & 24.8 & 26.1 & 26.7 & 23.3 \\

    \hline
    {\tt 3D~}BiLSTM guided by BiLSTM & 14.3 & 23.3 & 23.1 & 23.8 & 22.0 & 21.3 \\
    {\tt 3E~}4$\times$posterior fusion of {\tt 3D~}& 11.7 & 20.2 & 19.2 & 19.7 & 18.5 & 17.9 \\
    {\tt 3F~}4$\times$ROVER of {\tt 3D~}& 13.0 & 20.6 & 20.9 & 21.2 & 19.9 & 19.1 \\

    \hline
    {\tt 3G~}BiLSTM distilled from 4$\times$posterior fusion (\tt{3E})& 13.7 & 23.1 & 22.4 & 22.9 & 21.7 & 20.8 \\
    \thline
   \end{tabular*}
 \end{center}
 \vspace{-2mm}
\end{table*}

%% file: TABLE_COMPARISON.tex
\begin{table}[t]
 \begin{center}
  \caption{Comparison with published CTC-based direct acoustic-to-word models. [\%]}
  \vspace{-3mm}
  \label{tab:comparison}
   \begin{tabular*}{\columnwidth}{@{\extracolsep{\fill}}lccc}
    \thline
    ~ & Alignments & SWB & CH \\
    \thline
    {\tt 4A~}Word CTC~\cite{audhkhasi2017building}& N &14.6 & 23.6 \\
    {\tt 4B~}Hierarchical subword CTC~\cite{sanabria2018hierarchical}& N& 14.0 & 27.1 \\
    {\tt 4C~}Joint CTC-CE~\cite{yu2018multistage}& Y &13.0 & 23.4 \\
    \hline
    {\tt 4D~}Our best single model & N & 13.7 & 23.1 \\
    {\tt 4E~}Our best posterior fusion& N& 11.7 & 20.2 \\
    \thline
   \end{tabular*}
 \end{center}
\end{table}

%% file: TABLE_COVERAGE.tex
\begin{table}[t]
 \begin{center}
  \caption{Coverage ratio of posterior spikes. [\%]}
  \vspace{-3mm}
  \label{tab:coverage}
   \begin{tabular*}{\columnwidth}{@{\extracolsep{\fill}}lcc}
    \thline
    ~ & Training data & Test data\\
    \thline
    UniLSTM phone CTC            & ~ & ~ \\
    ~~~2 non-guided models     & 68.4 & 66.9 \\
    ~~~Guiding and guided models & 91.7 & 89.4 \\
    ~~~2 guided models         & 88.1 & 86.6 \\
    \thline
    BiLSTM word CTC              & ~ & ~ \\
    ~~~2 non-guided models     & 38.8 & 34.4 \\
    ~~~Guiding and guided models & 92.3 & 85.7 \\
    ~~~2 guided models         & 88.2 & 82.9 \\
    \hline
    \thline
   \end{tabular*}
 \end{center}
 \vspace{-3mm}
\end{table}